\documentclass{article}

\usepackage{xspace}
\newcommand{\sysname}{{\scshape Chai}\xspace}

\usepackage{microtype}
\usepackage{graphicx}
\usepackage[font=small, textfont=bf, labelfont=bf, aboveskip=0pt]{caption}
\usepackage{subcaption}
\usepackage{pgfplots}
\usepackage{booktabs} %
\usepackage{balance} 

\definecolor{pastelBlue}{HTML}{64A7E8}   %
\definecolor{pastelGreen}{HTML}{76C78F}  %
\definecolor{pastelOrange}{HTML}{FFB86C} %
\definecolor{pastelRed}{HTML}{FF7F7F}    %

\usepackage{hyperref}

\newcommand{\paraf}[1]{\noindent\textbf{#1}}

\usepackage[preprint]{icml2026}

\usepackage{amsmath}
\usepackage{amssymb}
\usepackage{mathtools}
\usepackage{amsthm}

\usepackage[capitalize,noabbrev]{cleveref}

\theoremstyle{plain}

\theoremstyle{definition}

\theoremstyle{remark}

\usepackage[textsize=tiny]{todonotes}

\icmltitlerunning{\sysname: CacHe Attention Inference for text2video}

\begin{document}

\twocolumn[
  \icmltitle{\sysname: CacHe Attention Inference for text2video}

  \icmlsetsymbol{equal}{*}

  \begin{icmlauthorlist}
    \icmlauthor{Joel Mathew Cherian}{yyy}
    \icmlauthor{Ashutosh Muralidhara Bharadwaj}{yyy}
    \icmlauthor{Vima Gupta}{yyy}
    \icmlauthor{Anand Padmanabha Iyer}{yyy}
  \end{icmlauthorlist}

  \icmlaffiliation{yyy}{Georgia Institute of Technology}

  \icmlcorrespondingauthor{Vima Gupta}{vgupta345@gatech.edu}

  \icmlkeywords{Machine Learning, ICML}

  \vskip 0.3in
]

\printAffiliationsAndNotice{}

\begin{abstract}
  Text-to-video diffusion models deliver impressive results but remain slow because of the sequential denoising of 3D latents. Existing approaches to speed up inference either require expensive model retraining or use heuristic-based step skipping, which struggles to maintain video quality as the number of denoising steps decreases. Our work, \sysname, aims to use cross-inference caching to reduce latency while maintaining video quality. We introduce Cache Attention as an effective method for attending to shared objects/scenes across cross-inference latents. This selective attention mechanism enables effective reuse of cached latents across semantically related prompts, yielding high cache hit rates. We show that it is possible to generate high-quality videos using Cache Attention with as few as 8 denoising steps. When integrated into the overall system, \sysname is \textbf{1.65x - 3.35x} faster than baseline OpenSora 1.2 while maintaining video quality.
\end{abstract}
\vspace{-7mm}
\section{Introduction}

Text-to-video diffusion models have rapidly emerged as a core generative primitive across industries such as entertainment, advertising, and education. Recent advances in large scale diffusion modeling and transformer-based architectures have enabled systems capable of generating long, temporally coherent videos with high visual fidelity \citep{peebles2023scalable, ho2022imagenvideo, chen2024videocrafter2, opensora}. This progress has led to the deployment of large proprietary systems such as Sora \cite{videoworldsimulators2024} and Veo \cite{GoogleVeo2024}, alongside open-source models including OpenSora \cite{opensora} and VideoCrafter2 \cite{chen2024videocrafter2}. Despite these advances, inference latency remains a fundamental bottleneck that limits large scale and interactive deployment of text-to-video systems.

The latency bottleneck arises from the sequential nature of Denoising Diffusion Probabilistic Models (DDPMs), which generate samples by iteratively denoising Gaussian noise over many timesteps \citep{ho2020ddpm, SongME21}. Compared to text-to-image diffusion, each denoising step in video diffusion is substantially more expensive due to the size of spatiotemporal (3D) latents \citep{ho2022imagenvideo, peebles2023scalable}. As a result, state-of-the-art video diffusion models typically rely on 30--50 denoising steps to achieve acceptable video quality, leading to high end-to-end latency \citep{opensora, chen2024videocrafter2}.

Prior work on accelerating diffusion inference broadly falls into training-based and training-free approaches. Due to the substantial retraining cost and tight coupling to specific architectures, retraining-based acceleration is often impractical in deployment scenarios. Training-free approaches instead focus on eliminating redundant computation during inference. A substantial body of work exploits the observation that not all denoising steps contribute equally to the final output \citep{wimbauer2024cache, so2024frdiff, selvaraju2024fora, ma2024learning, wang2024atedm}.

In the context of video diffusion, recent systems such as AdaCache \cite{adacache}, TeaCache \cite{Liu_2025_CVPR}, FasterCache \cite{lv2024fastercache}, and MagCache \cite{magcache} use feature-difference heuristics to skip or reuse intermediate denoising steps. We call these methods \textit{intra-inference caching} because they reuse latent representations from earlier denoising steps as inputs for later steps within a single inference run. These approaches typically skip mid-to-late denoising steps, where feature changes are small. However, early denoising steps introduce large structural changes to the video and cannot be skipped without severe degradation in motion consistency, object structure, and temporal coherence. As a result, intra-inference caching methods impose a fundamental limit on achievable latency without sacrificing video quality.

An alternative acceleration strategy explored in prior work in text-to-image diffusion is \textit{cross-inference caching}, where intermediate latents from previous generations are reused to accelerate future inference runs. NIRVANA demonstrates that reusing early denoising steps across semantically similar prompts can substantially reduce latency \citep{agarwal2024nirvana}. However, naively extending NIRVANA to video diffusion is ineffective. Video prompts are typically longer, more descriptive, and more diverse than image prompts, resulting in low prompt-level similarity and poor cache hit rates under realistic workloads. Consequently, cross-inference reuse does not translate directly to the text-to-video domain.

In this work, we explore a training-free acceleration strategy based on reusing diffusion latents across inference runs that share semantic structure at a finer granularity. We observe that while full video prompts may differ substantially, they often share common entities such as objects or scenes. Building on this observation, we present \sysname, a training-free system for accelerating text-to-video diffusion through entity-level cross-inference caching.

A key technical challenge is integrating cached latents into a new diffusion trajectory without introducing noise or artifacts. To address this, we introduce \textbf{Cache Attention}, a novel attention mechanism that selectively reuses entity-level information from cached latents. Rather than directly substituting latents, Cache Attention injects cached information into the attention computation, enabling selective reuse of relevant structure while preserving prompt-specific details.

Another important consideration is the number of denoising steps executed during accelerated inference. Prior work has shown that aggressively reducing the number of steps below 10 often leads to severe degradation in video quality for diffusion-based video models \citep{opensora, adacache, magcache, lv2024fastercache}. Our empirical evaluation shows that \sysname can generate high-quality videos on the OpenSora~1.2 backbone with as few as 8 denoising steps. This choice reflects an empirically observed balance between latency reduction and video quality preservation, rather than a fundamental limit of the approach.

\noindent Overall we make the following contributions
\begin{itemize}
    \item We present \sysname, a \textbf{training-free} system for accelerating text-to-video diffusion through cross-inference caching. \sysname achieves \textbf{1.65x--3.35x end-to-end speedup} over OpenSora 1.2 at 52--100\% cache hit rates while preserving video quality.
    \item We introduce \textbf{Cache Attention}, a novel attention mechanism that enables selective reuse of entity-level information from cached diffusion latents. Using Cache Attention, \sysname enables high-quality video generation at substantially reduced denoising steps. We find that the quality of videos generated with 8 denoising steps in \sysname matches the quality of 30-step inference on OpenSora 1.2.
    \item We demonstrate that \sysname is practical and scalable through a comprehensive system evaluation, showing \textbf{high cache hit rates ($>$80\%) under modest storage budgets (1--5\,GB)}, making cross-inference reuse viable as a training-free deployment strategy.
\end{itemize}

\section{Preliminaries}

\subsection{Diffusion Models for Text-to-Video}
Modern text-to-video systems are primarily built on DDPMs, which generate videos via an iterative process that denoises Gaussian noise toward the data distribution. Prior approaches relied on the 3D-UNet architecture and combined spatial and temporal convolutions. Recent work in large-scale text-to-video systems such as OpenSora, VideoCrafter2, and HunyuanVideo uses Diffusion Transformers(DiT) to better utilize tokenization, model depth, and long-range attention across both time and space domains ~\cite{opensora, chen2024videocrafter2, peebles2023scalable}.

Despite efforts such as DDIM~\cite{SongME21} and DPM-Solver~\cite{lu2022dpm} to reduce the number of diffusion steps via improved solvers, these methods still require sequential execution. Each timestep in the process depends on the output of the previous step. Current state-of-the-art architectures still require 30-50 steps for acceptable video quality. Hence, the primary bottleneck to the large-scale deployment of these systems is inference time.

\subsection{Intra-Inference Caching}

\begin{figure}[ht!]
      \centering
      \includegraphics[width=0.7\columnwidth]{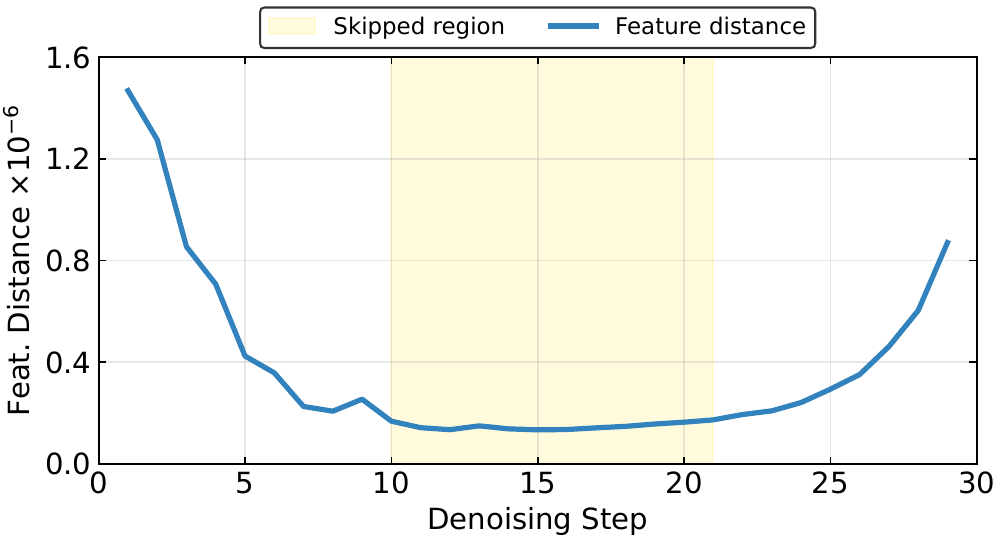}
      \caption{Feature distance between latents produced by adjacent denoising steps in a single text-to-video inference. The highlighted region indicates steps that are skipped
  by intra-inference caching approaches due to low degree of difference.}
      \label{fig:intra-inference-skipping}
  \end{figure}

A key observation across prior work is that each denoising step in the diffusion process does not contribute equally to the final output, as illustrated in Figure~\ref{fig:intra-inference-skipping}. Early steps in the process introduce large structural changes to the video, the middle-stage steps refine details, and the late-stage steps adjust textures and clean up residual noise. Intra-inference caching approaches ~\cite{adacache, Liu_2025_CVPR, magcache, selvaraju2024fora, so2024frdiff, wimbauer2024cache, ma2024learning} exploit this opportunity by skipping redundant steps using handcrafted heuristics. Among the various intra-inference caching approaches, we focus our comparison on AdaCache.

\subsubsection{AdaCache}

AdaCache~\cite{adacache} is an intra-inference caching method that leverages the feature distance between consecutive denoising steps to adaptively determine when to cache and when to forward a latent to a future step. This approach enables efficient skipping of intermediate steps, thus reducing inference time. In cases where the video has low motion or spatial complexity, fewer computations would be needed than for a video with high motion and complex feature changes. While AdaCache provides significant latency gains, its aggressive step-skipping policy results in a drastic loss of video quality.

\subsection{Cross-Inference Caching for Text-to-Video Diffusion Models}
Another acceleration approach limited to text-to-image diffusion models is cross-inference caching, which attempts to reduce diffusion latency by reusing the intermediate latent generated from a previous prompt with high similarity. The most dominant cross-inference text-to-image system is NIRVANA ~\cite{agarwal2024nirvana}. For every new prompt issued to NIRVANA, the system first performs a lookup within a vector database for a previous prompt with significant similarity. If a similar prompt is detected, the system searches the latent cache for entries associated with the matched prompt. For each prompt, NIRVANA stores five latent vectors from the 5th, 10th, 15th, 20th, and 25th denoising steps. The specific latent used for acceleration depends on the degree of similarity: less similar prompts utilize latents from earlier denoising steps, while more similar prompts use latents from later steps. By using a latent from the Kth step, the denoising process is able to skip the first K steps and thus reduce latency.

While this strategy is effective for image generation, naively extending NIRVANA to video is infeasible because video prompts are very dissimilar. This can be verified by analysing the VidProM \cite{wang2024vidprom} dataset, which is a collection of real-user prompts issued to   popular diffusion models. Figure~\ref{fig:opportunity-analysis} samples 3000 prompts from this dataset and maintains the first 100-1000 prompts in a cache, and then evaluates the similarity of the next 2000 prompts issued to the system. We can see that whole prompts show very little similarity with prompts that exist in the cache. Since NIRVANA's speedup is directly proportional to cache hits, low hit rates ($<$40\%) results in high latency.

\section{Methodology}

\subsection{Entity-Level Similarity}
Although whole-prompt similarity is low in text-to-video, a key empirical finding is that most video prompts share at least one object (e.g., tiger, zebra, car) or scene (e.g., forest, beach, kitchen). We refer to this type of similarity as \textit{Entity Similarity}. 

\begin{figure}[!ht]
      \centering
      \includegraphics[width=0.7\columnwidth]{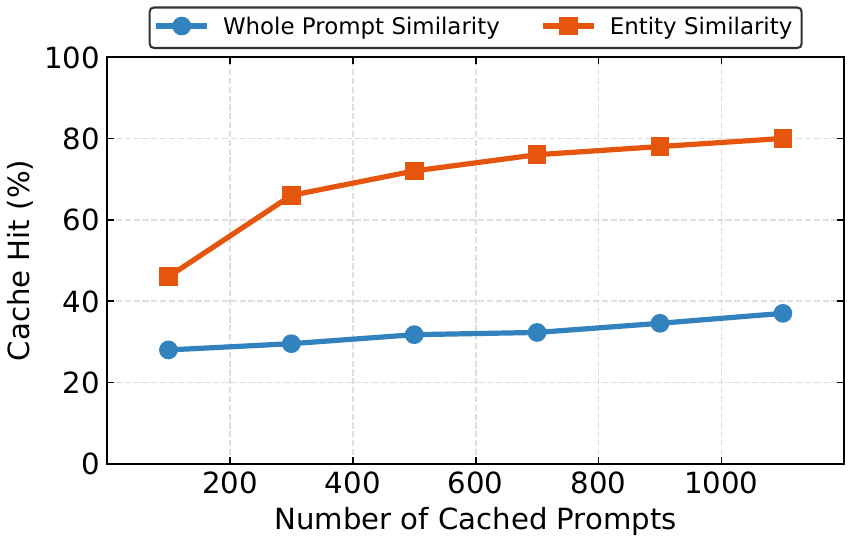}
      \caption{Cache hit rate (\%) vs.\ cache size on 2000 unseen VidProM prompts. Cached and unseen prompts show little overall similarity, but they share common entities and
  thus achieve a higher entity-similarity-based cache hit rate.}
      \label{fig:opportunity-analysis}
  \end{figure}

Figure ~\ref{fig:opportunity-analysis} shows that of 3000 prompts sampled from the VidProM \cite{wang2024vidprom} dataset, caching 300 prompts produces around 60\% cache hit rate, with cache hit rates increasing significantly to 80\% within realistic cache budgets. This suggests that there is scope to benefit from cross-inference reuse for text-to-video systems, but they hinge on exploiting caching at the granularity of objects/scenes.

\begin{figure*}[ht!]
    \centering
    \includegraphics[width=0.95\linewidth]{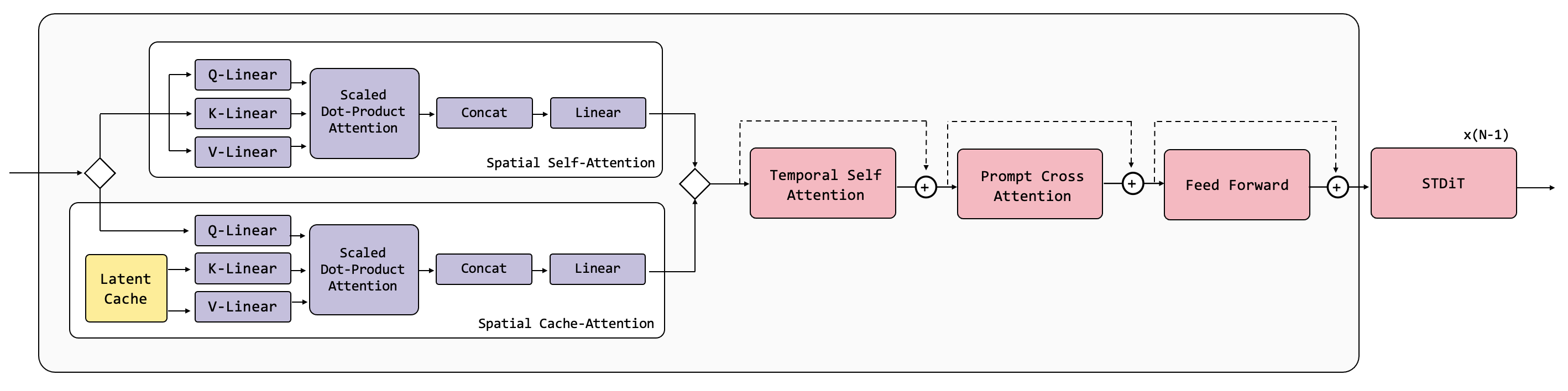}
    \caption{OpenSora STDiT block with the new Cache Attention layer capable of leveraging cross-inference entity reuse. When a cache hit occurs, the Cache Attention layer uses the latent cache as input to the key and value vectors to accelerate inference.}
    \label{fig:stdit-w-cache-attn}
\end{figure*}

\subsection{Cache Attention}
From the analysis in Figure~\ref{fig:opportunity-analysis}, it can be concluded that text-to-video cross-inference caching systems can only provide benefits when latents can be reused across prompts with similar entities (objects or background). These observations motivate Cache Attention, a mechanism that selectively attends to features of a cached latent that are relevant to the current prompt. \sysname is an entity-similarity-based cross-inference caching system that uses Cache Attention to reduce video diffusion latency.

When comparing feature distances between adjacent latents, we can observe that inferences with fewer denoising steps tend to make larger changes per step than those with more denoising steps. This is an outcome of the rectified flow scheduler~\cite{liu2022flow}, which determines the degree of change introduced with each denoising step. For example, during an 8-step inference, the model takes larger steps than during a 30-step run, where it takes smaller steps and converges to generate a video of better quality. Cache attention leverages this observation by using latents from the more focused 30-step run to help drive the diffusion process in the 8-step run.

\begin{figure}[ht]
    \centering
    \includegraphics[width=0.9\columnwidth]{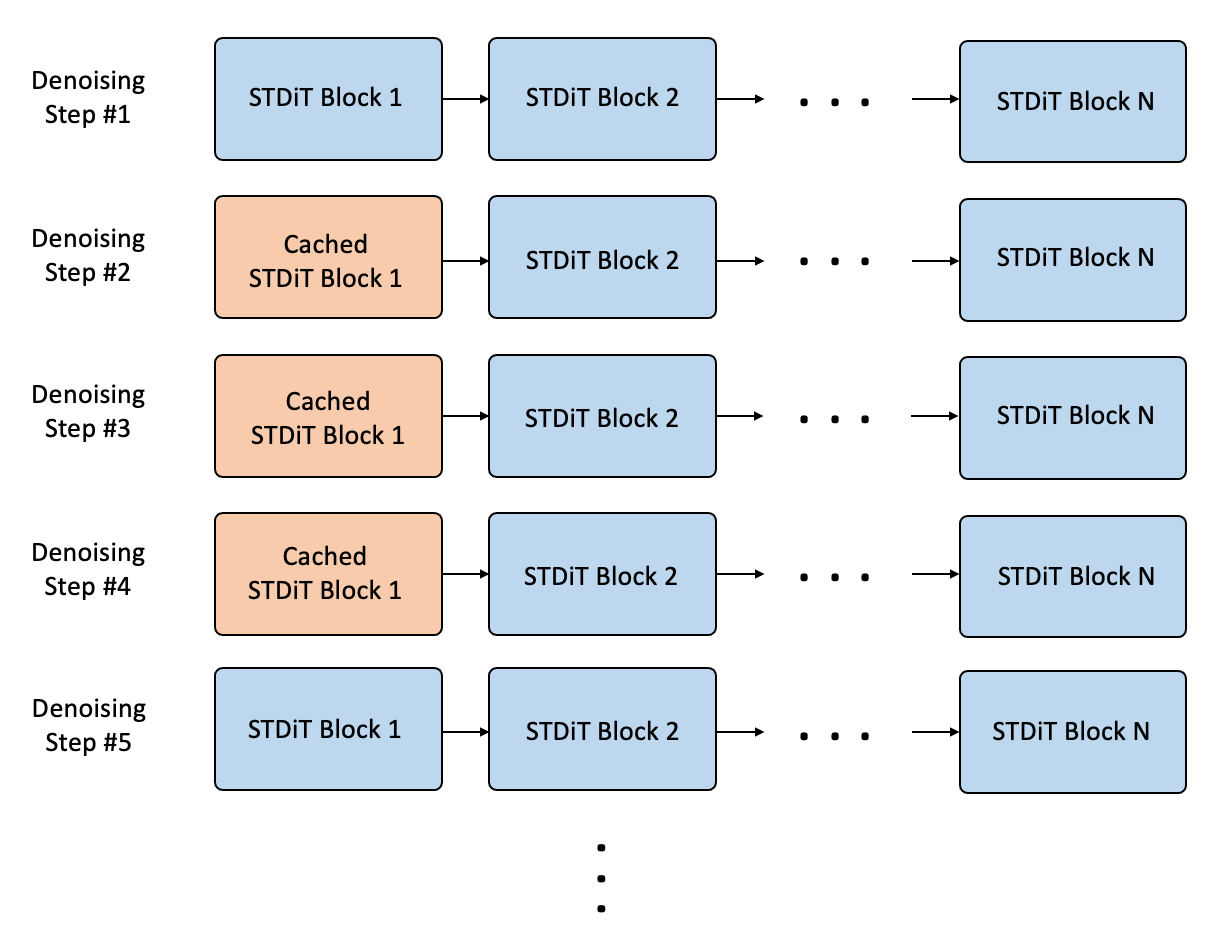}
    \caption{During cross-inference reuse, the STDiT blocks are scheduled to use the latent cache for the first block in the 2nd, 3rd, and 4th denoising steps.}
    \label{fig:cache-schedule}
    \vspace{-2mm}
\end{figure}

Figure~\ref{fig:stdit-w-cache-attn} depicts the modified STDiT block from OpenSora1.2 with the new Cache Attention Layer. The layer replaces the previous Spatial Self Attention layer. The main difference between the two layers lies in the key and value inputs to the attention operation. In the case of Cache Attention, the key and value vectors are derived from a cached latent space that shares similar entities with those specified in the input prompt. The use of latent cache for only the key and value entries is because it allows the prompt-modulated Gaussian noise query (Q) to selectively pick features of relevant entities from the latent cache rather than using the entire latent. 

Our experiments show that cached latents can be scheduled as shown in Figure \ref{fig:cache-schedule} for optimal performance. We intentionally avoid using the cached block in the first denoising step because the query vector for that block is just random Gaussian noise rather than prompt-modulated Gaussian noise for every other denoising step. For the remaining denoising steps, we can match the latents from the same denoising step in the cache directly to the relevant block. Our experimentation has shown that caching provides little benefit after the 4th denoising step. Since cache usage for more denoising steps requires more storage, we limit it to the 2nd, 3rd, and 4th denoising steps. We also observe that cache usage across successive blocks within the same denoising step can actually cause the model to insert more noise; hence, we limit cache usage to the first block in each denoising step.

\begin{figure*}[ht]
    \centering
    \includegraphics[width=\linewidth]{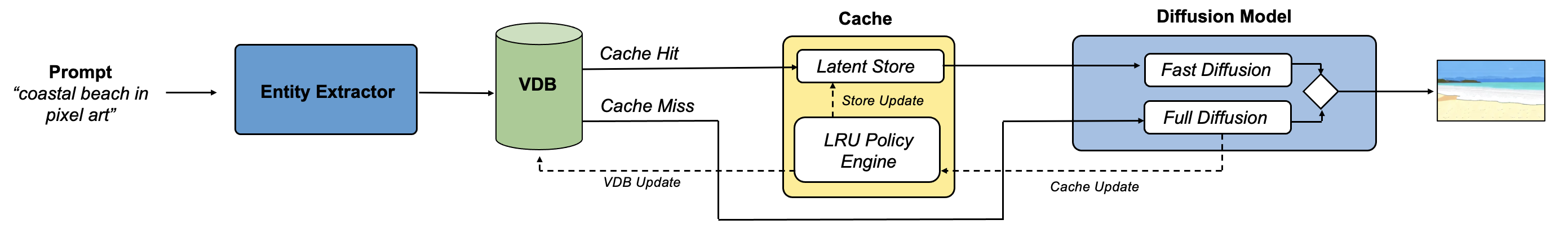}
    \caption{\sysname design. Each new prompt is compared against previous ones. On a cache hit, it retrieves and reuses stored latents to enable faster inference with fewer denoising steps. On a cache miss, it performs full inference and caches new latents for future use. The cache policy engine manages storage by evicting older latents once the cache exceeds a set limit.}
    \label{fig:chai2-design}
\end{figure*}

\begin{figure}[ht!]
    \begin{subfigure}[t]{0.45\textwidth}
        \includegraphics[width=1.05\linewidth,height=5cm]{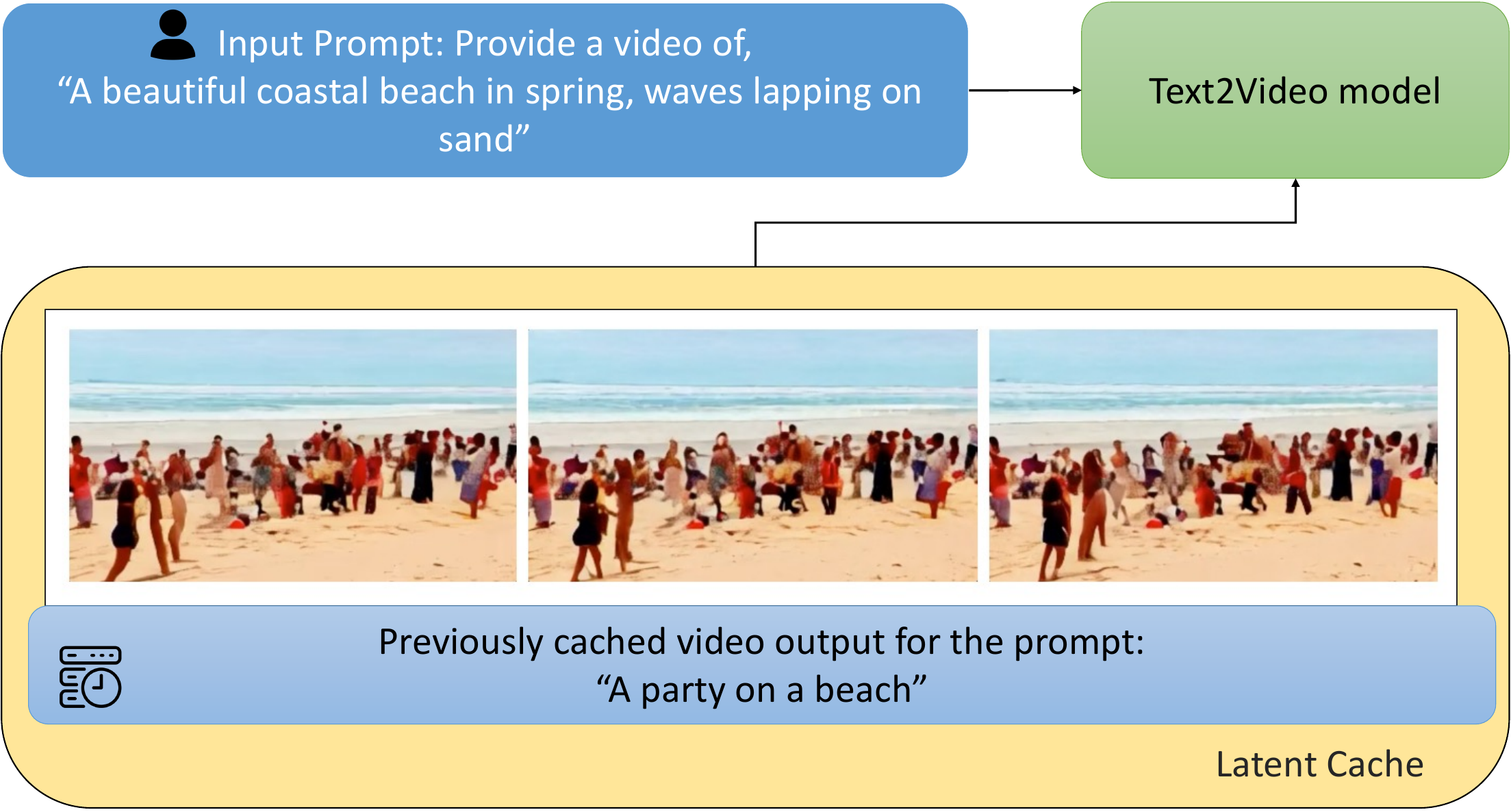}
        \caption{Cached video (30 steps)}
    \end{subfigure}
    \\
    \begin{subfigure}[t]{0.45\textwidth}
        \includegraphics[width=1.05\linewidth,height=2.5cm]{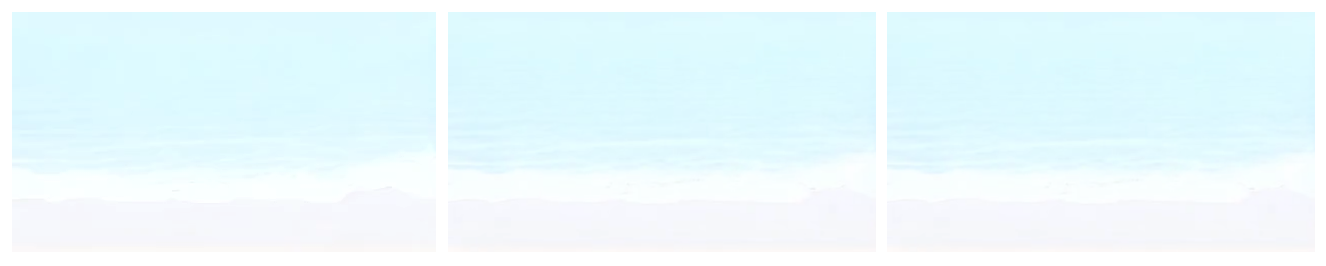}
        \caption{OpenSora 1.2 (8 steps)}
    \end{subfigure}
    \\
    \begin{subfigure}[t]{0.45\textwidth}
        \includegraphics[width=1.05\linewidth,height=2.5cm]{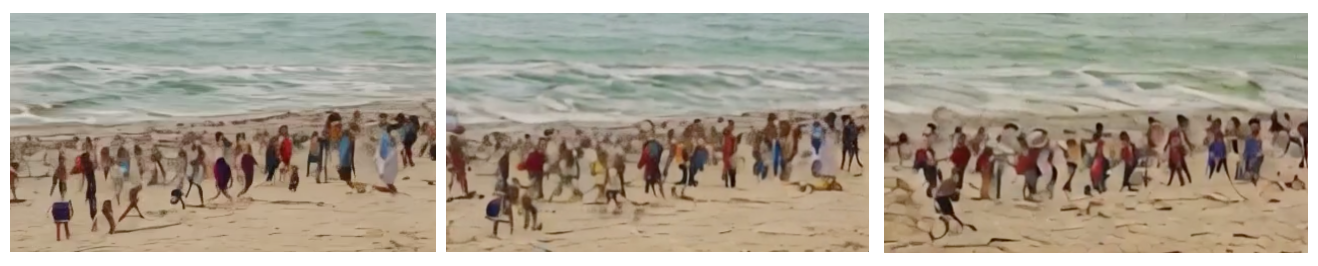}
        \caption{NIRVANA-VID (8 steps)}
    \end{subfigure}
    \\
    \begin{subfigure}[t]{0.45\textwidth}
        \includegraphics[width=1.05\linewidth,height=2.5cm]{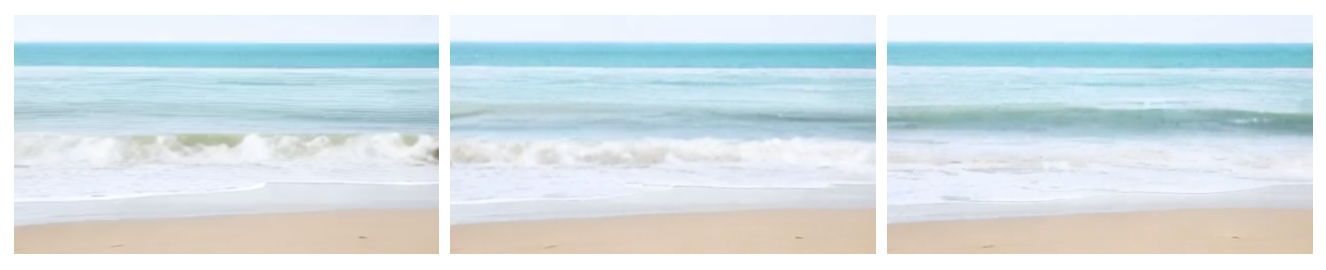}
        \caption{\sysname (8 steps)}
    \end{subfigure}
    \caption{Videos generated by various text-to-video systems for the prompt \textit{“A beautiful coastal beach in spring, waves lapping on sand”} when run for 8 denoising steps. NIRVANA-VID and \sysname reuse cached latents from a similar prompt: \textit{”A party on a beach”}. The video produced by \sysname has the least amount of noise and adheres more closely to the prompt.}
    \label{fig:comparison-of-methods}
    \vspace{-4mm}
\end{figure}

While the evaluation section discusses the quality of videos generated with Cache Attention, Figure~\ref{fig:comparison-of-methods} illustrates how this technique compares with other systems. We consider the case where a video needs to be generated for the prompt \textit{"A beautiful coastal beach in spring, waves lapping on sand"}. 

To compare results, we run all systems for 8 steps. We can see that in 8 steps, OpenSora 1.2 is only capable of producing boundaries where eventually a beach could develop (Figure~\ref{fig:comparison-of-methods}(b)). NIRVANA-VID and \sysname use cached latents generated for the prompt \textit{"A party on a beach"}, which has a cosine similarity of 0.58 with the prompt in consideration. It can be seen that NIRVANA-VID (Figure ~\ref{fig:comparison-of-methods} (c)) does better than OpenSora 1.2, but the video is noisy and fails to separate out the "beach" aspect of the prompt from the "party" component. Figure ~\ref{fig:comparison-of-methods}(d) shows that in 8 steps \texttt{Chai} managed to selectively pick out the beach component from the cache and generate a video that adheres to the prompt with minimum noise.

\subsection{\sysname Design}

The design of \sysname is illustrated in Figure~\ref{fig:chai2-design}. The system consists of several key components that work together to enable entity-level cross-inference caching.

When a prompt is issued to the system, an \textit{Entity Extractor} identifies the entities present in the prompt, including objects and background elements. These extracted entities enable the identification of reusable semantic structures across different inference runs.

The system maintains a \textit{Vector DB} that stores embeddings associated with entities encountered in previous prompts, along with metadata that maps each embedding to its corresponding cached latent.

Cached diffusion latents themselves are stored in a \textit{Latent Store}, which maintains a collection of latents associated with previously processed prompts. These latents are reused during accelerated inference when matching entities are identified.

Cache capacity is managed by an \textit{LRU Policy Engine}, which tracks the last access time of cached latents. All cache writes pass through this unit. When a cache write would cause the cache size to exceed a predefined threshold, the policy engine identifies the least recently used cache item, evicts it from the latent store, and removes the associated indexes from the vector database.

Finally, \sysname operates on top of a DiT-based \textit{Diffusion Model}. In this work, we build on OpenSora~1.2 and modify the model to support two diffusion modes: \textit{full} and \textit{fast}. The full mode is triggered on a cache miss and executes all denoising steps using vanilla self-attention, with cache attention layers disabled. The fast diffusion mode is triggered on a cache hit, runs for a reduced number of denoising steps, and enables cache attention layers to incorporate information from cached latents.

\section{Evaluation}

We perform comprehensive evaluation of our technique \sysname with the goal of answering four research questions:
\begin{enumerate}
    \item Does Cache Attention maintain video quality while reducing the number of denoising steps and overall latency?
    \item How does Cache Attention perform under constrained cache budgets, and how does that impact the latency benefits provided by \sysname?
    \item How does \sysname scale with varying cache budgets?
    \item How does cross-inference caching in CHAI compare to intra-inference acceleration methods?
\end{enumerate}

\paraf{Baselines} To evaluate the performance of our proposed method, we compare \sysname which runs on 8 denoising stpes, chosen based on an empirical latency and quality tradeoff against the following systems: \textit{OpenSora~1.2}, the baseline OpenSora model configured to run for 30 denoising steps; \textit{NIRVANA-VID}, a strawman baseline that adapts NIRVANA-style whole-prompt reuse to the video domain; and \textit{AdaCache}, an intra-inference caching system that executes a heuristic-driven number of denoising steps. All systems are configured to generate a 3-second 240px video and are evaluated on a single NVIDIA H100 GPU for consistency.

\paraf{Datasets} Our evaluation is based on two workloads, namely VBench and VidProM. 

The \textbf{VBench Test Suite} is a state-of-the-art, standardized evaluation benchmark for text-to-video diffusion models, comprising prompts spanning diverse motion patterns, scene compositions, temporal behaviours, and object categories. We evaluate 692 prompts across 12 VBench dimensions, including background consistency, temporal flickering, motion smoothness, aesthetic quality, imaging quality, object class, color, spatial relationship, scene, appearance style, temporal style, and overall consistency, with a single video generated per prompt. VBench computes a weighted aggregate score, which we use as our overall video-quality metric. 

To evaluate cross-inference reuse under realistic prompt diversity, we use 1000 prompts from \textbf{VidProM}, a dataset of real user prompt inputs to video diffusion models. Because custom prompts can only be evaluated on 6 VBench dimensions, we limit the reported VidProM VBench scores to these dimensions.

\paraf{Implementation}  \sysname is built on top of OpenSora~1.2. It leverages \texttt{Spacy} 3.8.11 for part-of-speech tagging to extract entities from prompts. The prompts are converted to embeddings using the LongCLIP model \cite{zhang2024longclip}. \sysname also uses Faiss 1.13 as the vector database for efficient similarity search over entity embeddings.

\paraf{Evaluation Metrics} We measure four metrics: end-to-end video generation latency, the number of denoising steps executed, VBench score, and cache hit rate.

\subsection{Video Quality and Latency Evaluation}

To evaluate the effectiveness of Cache Attention, we benchmark \sysname, NIRVANA-VID, and OpenSora 1.2 on the VBench dataset. Ahead of the evaluation, we populate the \sysname cache with entity-similar prompts to ensure that all inferences trigger the \textit{fast diffusion} path with the cache attention layer enabled. Similarly, we populate the NIRVANA-VID cache with similar prompts, but due to limitations imposed by whole-prompt matching, only 75\% of the prompts reuse cached latents.
\begin{figure}[!ht]
      \centering
      \includegraphics[width=0.8\columnwidth]{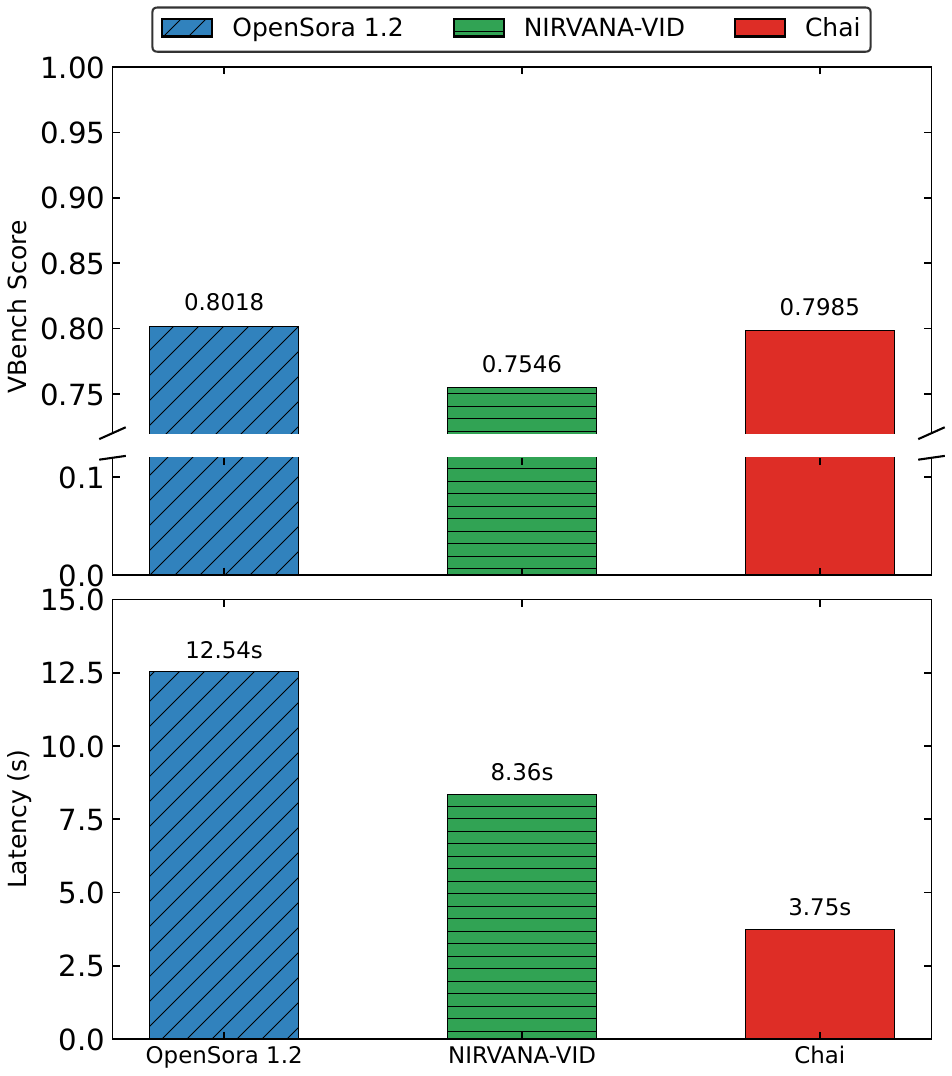}
      \caption{VBench quality and latency comparison on 692 prompts.
      \sysname achieves highest speedup of 3.35$\times$ while maintaining
      near-baseline video quality (0.4\%).}
      \label{fig:vbench-quality}
  \end{figure}

In Figure~\ref{fig:vbench-quality}, we observe that OpenSora 1.2 achieves the highest VBench score (0.8018) but also the highest latency since it executes all 30 denoising steps. NIRVANA-VID reduces latency by reusing latents to skip early denoising steps, but suffers from severe quality degradation because its direct latent-substitution approach prevents it from masking unnecessary entities from the cached latent. On the other hand, with only 8 denoising steps (3.75$\times$ lower), \sysname achieves a VBench score of 0.7985 (0.3\% below the baseline) and a 3.35$\times$ speedup over OpenSora 1.2. Additionally, \sysname outperforms NIRVANA-VID in terms of quality while maintaining lower latency.

These results validate the claim that \textbf{Cache Attention can preserve quality while substantially reducing the number of denoising steps}.

\subsection{Performance under constrained scenarios}

\begin{figure}[!ht]
      \centering
      \includegraphics[width=0.9\columnwidth]{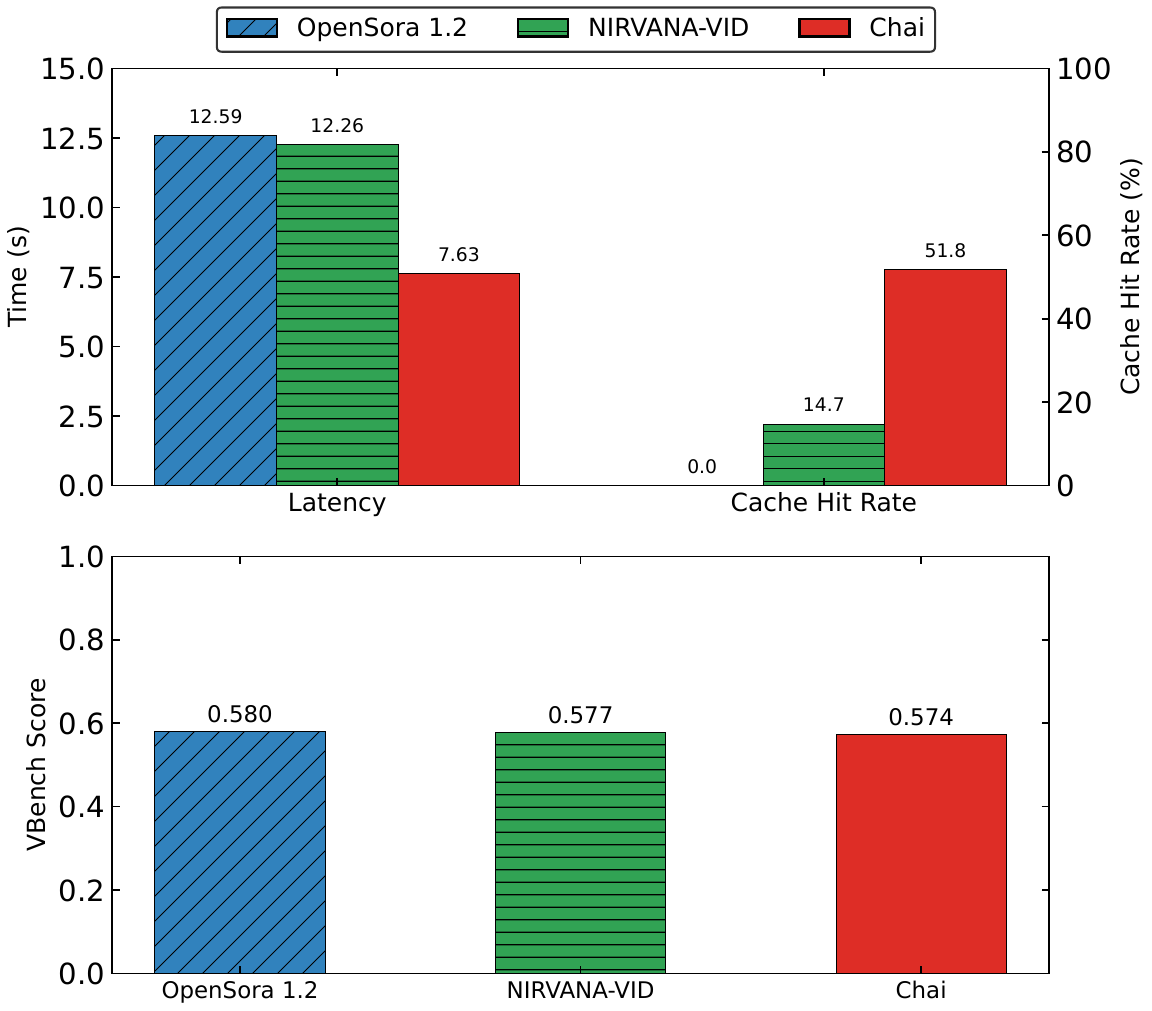}
      \caption{Cross-inference caching on the VidProM with a
      constrained cache budget. \sysname achieves 1.65$\times$ speedup over Open Sora 1.2 while maintaining comparable video
      quality.}
      \label{fig:vidprom-multi}
  \end{figure}

In Figure~\ref{fig:vidprom-multi}, we  showcase the behavior of \sysname as a cross-inference caching system under tighter cache budgets to closely resemble real-world scenarios. In this evaluation, we use the VidProM dataset (sorted by date) to simulate prompts issued to a diffusion model system in a real-world scenario. We populate our cache with the first 100 prompts and evaluate the system's performance on the following 1000 prompts. 

Under a 10\% cache size constraint, whole-prompt reuse performs poorly. VidProM prompts are long, descriptive, and rarely repeated, resulting in low whole-prompt hit rates (14.7\%). As a result, NIRVANA-VID provides almost no latency improvement over OpenSora 1.2 (12.26\,s vs.\ 12.59\,s).

\textbf{Entity-level reuse, in contrast, remains effective even with a small cache.} \sysname retrieves cached latents using object and scene embeddings, substantially improving hit rate (52\%). With the same 10\% cache, \sysname reduces latency to 7.63\,s ($1.65\times$ faster than OpenSora 1.2) with a limited drop in VBench score.

This result shows that entity-level similarity is significantly more robust than whole-prompt matching, especially in the diverse prompt distributions typical of production workloads. Additionally, we show that the \sysname system can reduce latency with minimal degradation in video quality.

\subsection{Storage Overhead and Cache Management}

Cross-inference caching is only helpful if the cache is small enough to fit in memory while still achieving high hit rates. In \sysname, each cached prompt consists of three latent tensors, totaling about 1.3 MB (for 240px video latents on the OpenSora 1.2 backbone).

\begin{figure}[!ht]
      \centering
      \includegraphics[width=\columnwidth]{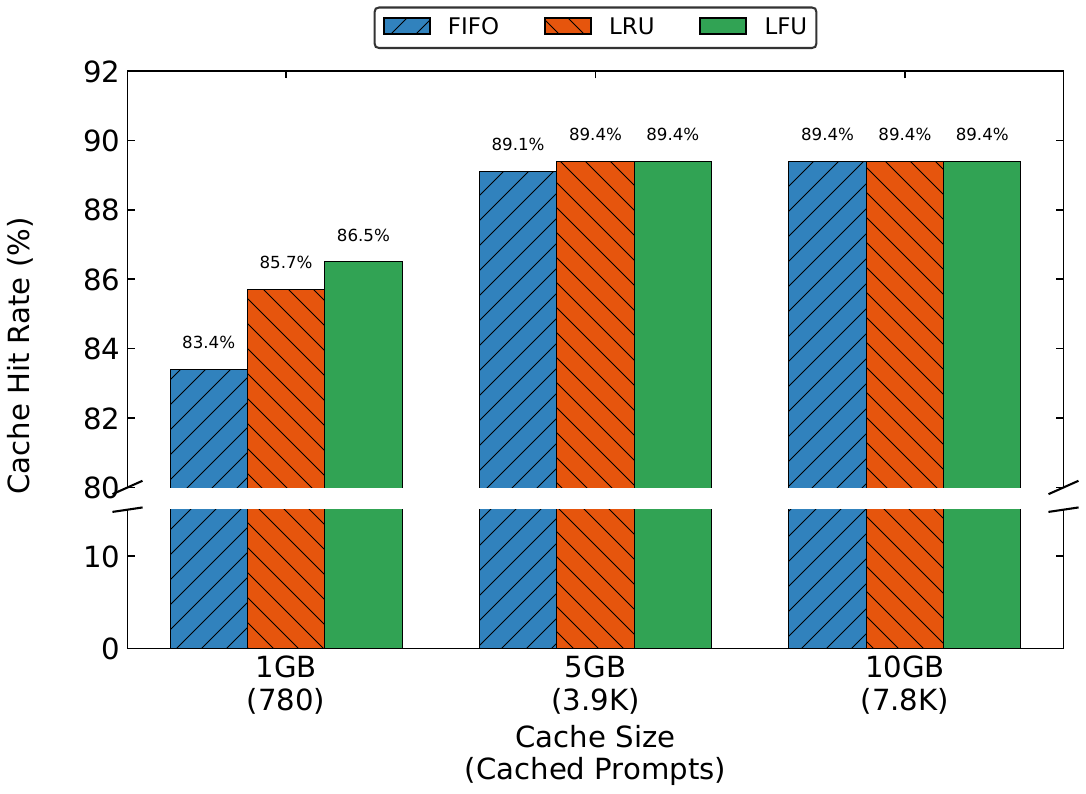}
      \caption{Cache hit rate scaling with increasing cache budget under
      different replacement policies. \sysname achieves over 83\% hit
      rate with just 1GB of cache (780 prompts), quickly saturating to
      89.4\% at 5GB, demonstrating
      that \sysname's entity-based caching is effective even under
      tight memory constraints.}
      \label{fig:cache-scaling}
  \end{figure}

Figure~\ref{fig:cache-scaling} reports cache-hit rates under different cache size budgets for various cache replacement policies. The evaluation is performed over 100,000 prompts from the VidProM dataset. 
The cache replacement policies under consideration include \textit{FIFO}, which evicts the oldest item in the cache; \textit{LRU}, which evicts the least recently accessed item; and \textit{LFU}, which evicts the least frequently accessed item.

\textbf{The results suggest that even modest cache sizes (1--5\,GB) yield high hit rates.} Additionally, LRU and LFU provide the best cache hit rates for smaller cache sizes, but the hit rates for all policies even out for larger caches.

\subsection{Intra-Inference vs.\ Cross-Inference Caching}
\label{subsec:intra-vs-cross}
\begin{figure}[!ht]
      \centering
      \includegraphics[width=0.8\columnwidth]{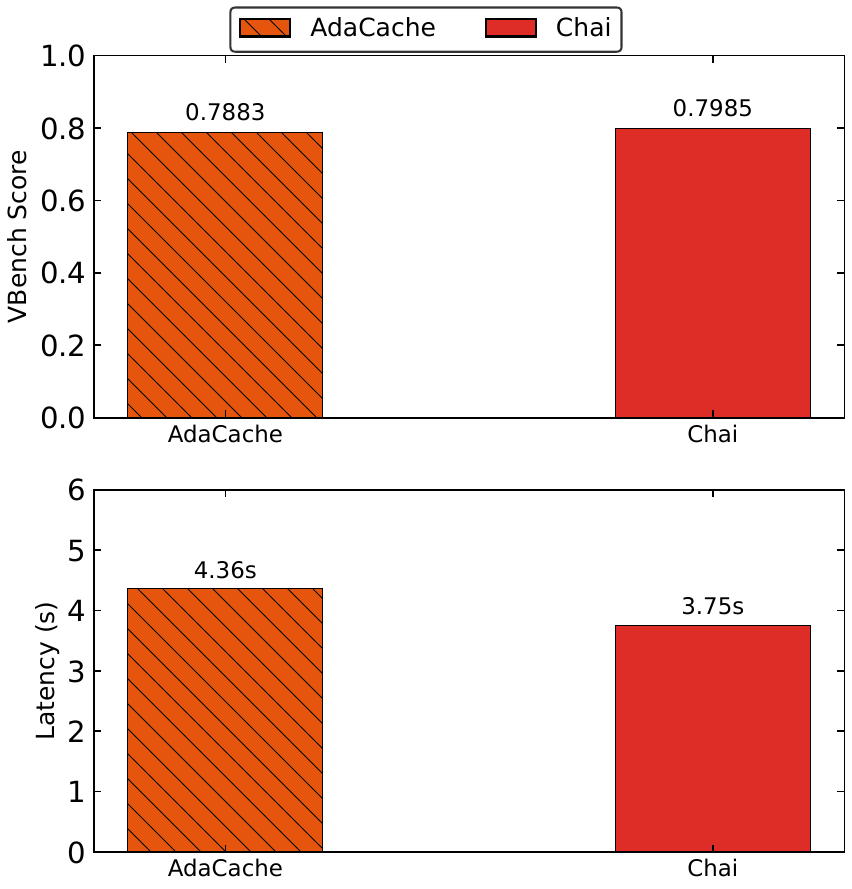}
      \caption{On the Vbench benchmark, \sysname produces videos with better quality while maintaining lower latency than AdaCache.}
      \label{fig:cache-scaling2}
  \end{figure}

Figure~\ref{fig:cache-scaling2} summarizes the quality and latency behavior of \sysname and AdaCache on the VBench workload. AdaCache reduces latency by skipping mid- and late-stage denoising steps within a single inference, achieving an average VBench score of 0.7883 with an average latency of 4.36\,s. In contrast, \sysname operates with 8 denoising steps under fast diffusion and achieves a higher VBench score of 0.7985 at a lower latency of 3.75\,s.

This result indicates that \sysname's cross-inference caching can exceed the quality and latency limits imposed by intra-inference step skipping alone. However, AdaCache and related methods remain effective when no reusable cache entry exists. 

\section{Related Work}

\subsection{Training-Based Acceleration}
DC-VideoGen~\cite{li2024dcvideogen} proposes a training-based approach to accelerating inference by transferring weights from a pretrained video diffusion model into a new architecture with a more compressed latent space. This transition relies on an adaptive alignment mechanism that continuously aligns representations between the source and target models. While DC-VideoGen demonstrates significant speedups, it incurs substantial retraining costs. The required fine-tuning takes approximately 10 days, even when starting from a strong base model such as Wan2.1, making such approaches impractical for scenarios requiring rapid deployment or frequent model updates.

\subsection{Training-Free Intra-Inference Caching}
AdaCache, TeaCache, MagCache, PAB, TaoCache, and related methods reduce redundant computation within a single inference by skipping denoising steps where residual changes fall below a threshold. These methods improve mid- to late-stage denoising efficiency, but share a fundamental limitation: they cannot skip early diffusion steps where feature changes are significant. As a result, a fixed latency overhead remains unavoidable with these techniques.

\subsection{Cross-Inference Reuse}
NIRVANA~\cite{agarwal2024nirvana} demonstrates that early denoising steps can be reused across image prompts that have sufficient similarity. However, a naive extension of NIRVANA to text-to-video diffusion is ineffective due to the long, descriptive, and diverse nature of video prompts, resulting in low cache hit rates under whole-prompt similarity matching. As a result, prior cross-inference reuse strategies do not translate effectively from image to video generation.

\subsection{Miscellaneous Acceleration Strategies}
Step reduction strategies aim to minimize the number of denoising iterations. For example, progressive distillation~\cite{salimans2022distillation} trains a student model to approximate longer sampling trajectories with fewer steps. While effective, these methods typically require retraining or exhibit sensitivity to sampling noise.

Reverse Transition Kernel (RTK)~\cite{zhang2024rtk} approaches acceleration from a theoretical perspective by formulating diffusion inference as a sequence of log-concave sampling subproblems solvable using MALA or ULD, offering improved convergence guarantees over DDPM and DDIM.

Latency reduction at the architectural level is another active area of research. AT-EDM~\cite{wang2024atedm} and Token Merging~\cite{bolya2023token} apply token pruning or compression by identifying and discarding less important latent tokens during inference. SnapFusion~\cite{li2023snapfusion} demonstrates efficient text-to-image generation on mobile devices through aggressive model compression and optimization. Similarly, Parallel Diffusion~\cite{shih2024parallel} reformulates diffusion sampling to enable concurrent execution of denoising steps.

\noindent \sysname targets a design point that prior work does not address. It adopts a training-free approach based on cross-inference latent reuse. Unlike NIRVANA, \sysname operates on entity-level similarities and integrates cached information through attention mechanisms rather than direct latent substitution. Compared with intra-inference methods, Chai surpasses the limit of skippable steps while maintaining video quality. 

\section{Future Work}

While \sysname demonstrates that cross-inference reuse can substantially reduce denoising latency with minimal quality loss, several directions remain for future exploration.

\textbf{Efficient Latent Storage:}
Our current design stores a small number of cached latents per prompt, but higher-resolution video generation will increase memory demands. Exploring lightweight latent compression techniques, such as quantization-aware storage, could further reduce cache footprint while preserving the structural information required for Cache Attention.

\textbf{Richer Cache Utilization:}
\sysname currently reuses a single cached latent per inference. Extending Cache Attention to selectively combine information from multiple cached latents that share overlapping entities or scenes may improve robustness under diverse prompts, but requires careful handling to avoid interference between reused features.

\textbf{Joint Intra- and Cross-Inference Acceleration:}
While this work focuses solely on cross-inference, a natural next step is to design a unified inference scheduler that combines both intra- and cross-inference caching within a single generation. Such a system could adaptively decide when to reuse cached latents across runs and when to skip or reuse steps within an inference, based on prompt semantics and denoising-stage sensitivity.

\section{Conclusion}

In this work, we presented \sysname, a training-free cross-inference caching system for accelerating text-to-video diffusion models. By introducing \emph{Cache Attention}, \sysname selectively reuses entity-level information from prior inferences, enabling effective cross-inference reuse without degrading video quality. Across two evaluation workloads, \sysname demonstrates that entity-level caching is practical and effective for video diffusion: it maintains video quality within $0.3\%$ of baseline OpenSora~1.2 while achieving a $3.35\times$ speedup on VBench using just 8 denoising steps, and delivers a $1.65\times$ speedup on the diverse VidProM workload where whole-prompt reuse is ineffective. Furthermore, \sysname achieves high cache hit rates even under modest cache budgets, making selective cross-inference latent reuse a scalable and deployment-ready acceleration strategy.

\balance
\bibliography{ref}
\bibliographystyle{icml2026}

\end{document}